\documentclass[format=sigconf,nonacm]{acmart}

\usepackage{tikz,adjustbox,subfig,nicefrac}
\usepackage{amsthm,amsmath,verbatim}

\usetikzlibrary{calc,arrows}

\newcommand{\tp}{{\mbox{\tiny \sf T}}}
\newcommand{\chd}{\mathcal C}
\renewcommand{\Re}{\mathbb{R}}

\newcommand{\T}{\mathcal T}
\newcommand{\W}{\mathcal W}
\newcommand{\Nint}{\mathcal N_{\text{int}}}
\newcommand{\Nlf}{\mathcal N_{\text{leaf}}}
\newcommand{\N}{\mathcal N}
\newcommand{\Q}{\mathcal Q}

\newcommand{\z}{\mathbf z}

\newtheorem{problem}{Problem}

\title[Information-Theoretic Abstractions via Mixed-Integer Linear Programming]{Information-Theoretic Abstractions for Resource-Constrained Agents via Mixed-Integer Linear Programming}

\author{Daniel T. Larsson}
\orcid{}
\affiliation{%
	\position{PhD Student}
	\institution{Georgia Institute of Technology}
	\department{School of Aerospace Engineering}
	\city{Atlanta}
	\state{GA}
	\country{USA}}
\email{daniel.larsson@gatech.edu}

\author{Dipankar Maity}
\orcid{}
\affiliation{%
	\position{Assistant Professor}
	\institution{University of North Carolina}
	\department{Department of Electrical and Computer Engineering}
	\city{Charlotte}
	\state{NC}
	\country{USA}}
\email{dmaity@uncc.edu}

\author{Panagiotis Tsiotras}
\orcid{}
\affiliation{%
	\position{Andrew and Lewis Chair Professor}
	\institution{Georgia Institute of Technology}
	\department{School of Aerospace Engineering}
	\department{Institute for Robotics and Intelligent Machines}
	\city{Atlanta}
	\state{GA}
	\country{USA}}
\email{tsiotras@gatech.edu}

\begin{document}

\begin{abstract}
In this paper, a mixed-integer linear programming formulation for the problem of obtaining task-relevant, multi-resolution, graph abstractions for resource-constrained agents is presented.
The formulation leverages concepts from information-theoretic signal compression, specifically the information bottleneck (IB) method, to pose a graph abstraction problem as an optimal encoder search over the space of multi-resolution trees.
The abstractions emerge in a task-relevant manner as a function of agent information-processing constraints, and are not provided to the system a priori.
We detail our formulation and show how the problem can be realized as an integer linear program.
A non-trivial numerical example is presented to demonstrate the utility in employing our approach to obtain hierarchical tree abstractions for resource-limited agents.
\end{abstract}

\maketitle

\section{Introduction}

The preservation of task-relevant information is considered central to the formation of abstractions for autonomous systems \cite{Ponsen2010}.
The process of obtaining abstractions has traditionally been handled by system designers, and is seldom left for the agents to discover on their own \cite{Zucker2003,Ponsen2010}.
However, it has long been considered crucial to solving complex problems that autonomous agents themselves have the ability to identify task-relevant structures contained in data \cite{Zucker2003}.
Determining the relevant aspects of data allows agents to form abstractions by ignoring irrelevant details, thereby reducing the computational complexity of problem solving \cite{Zucker2003,Ponsen2010}.

In the robotics community, a number of studies have leveraged abstractions to reduce the computational complexity of problem solving.
For example, the authors of \cite{Tsiotras2011} develop a framework for path-planning in dynamic environments that employs abstractions in order to reduce the execution time of graph search algorithms such as A\(^*\) and Dijkstra.
To do so, the framework presented in \cite{Tsiotras2011} utilizes wavelets to form environment abstractions, where the relevant region of the world is considered to be that closest to the agent.
Consequently, the approach developed in \cite{Tsiotras2011} reduces the computational complexity by executing the graph search problem on a reduced graph of the environment, which is formed by aggregating vertices in the original space that are considered distant from the agent.
To balance path optimality with computational complexity, the authors of \cite{Tsiotras2011} recursively resolve the planning problem as the agent moves through the environment and reasons about the world.

Similarly to \cite{Tsiotras2011}, the authors of \cite{Hauer2015,Kambhampati1986} leverage environment abstractions in the form of variable-depth hierarchical (quad)tree representations in order to reduce the computational complexity of path-planning.
The use of hierarchical tree structures allows the work of \cite{Hauer2015} to develop an approach utilizing probabilistic trees, thereby enabling the incorporation of inherent sensor uncertainties present when autonomous agents explore unknown maps \cite{Kraetzschmar2004,Hornung2013}.
Employing probabilistic tools to model environment uncertainty allows the framework of \cite{Hauer2015} to be used in conjunction with occupancy grid (OGs) representations of the operating space, which can be updated as the agent traverses the world \cite{Thrun2006}.
However, existing works rely on the selection of tuning parameters and do not provide a method to obtain abstractions explicitly in terms of the agent's potential resource constraints.

The essence of abstraction, that is, the removal of irrelevant information, is not unique to the robotics or autonomous systems communities.
Information theorists have considered the abstraction problem when developing frameworks for signal compression in capacity-limited communication systems \cite{Cover2006, Tishby1999}.
For example, the framework of rate-distortion (RD) theory approaches signal compression by formulating an optimization problem so as to maximize compression, measured by mutual information between the compressed and the original signal, subject to constraints on the quality of the reduced representation \cite{Cover2006}.
The quality of the compressed representation is measured by a predefined distortion function, which implicitly identifies which features in the original signal are important, or relevant, in order for the reduced representation to achieve low distortion.
A related approach is that of the information bottleneck (IB) method, which replaces the distortion function with the mutual information between the compressed representation and a relevant auxiliary variable \cite{Tishby1999}.
In this way, the IB method formulates an optimal encoder problem where the goal is to maximize compression while retaining as much information as possible regarding the auxiliary variable.
Owing to its general statistical formulation, the IB method has been applied to problems ranging from categorical classification of manuscript data to latent space representation using deep learning \cite{Slonim2000,Alemi2016,Kolchinsky2017,Lu2020a}.

In this paper, we model agent resource constraints as limitations to the system's information-processing capabilities \cite{Lipman1995,Genewein2015}.
Accordingly, we develop an information-theoretic approach towards obtaining multi-resolution environment abstractions that can be tailored to agent resource constraints.
Our approach utilizes the information bottleneck principle to formulate an information-theoretic, task-relevant, graph abstraction problem that maximizes compression while limiting the amount of relevant information contained in the resulting abstraction.
The amount of retained relevant information can be adjusted to suit the systems information-processing capabilities.
Furthermore, the abstractions are emergent in the form of multi-resolution (quad)trees that are not provided to the system a priori.
We show how our problem can be re-cast as a mixed-integer linear program (MILP), and demonstrate the utility of the approach on a non-trivial numerical example.

\section{Problem Formulation}

\begin{figure}
	\begin{adjustbox}{max size={0.46\textwidth}}
	\begin{tikzpicture}[scale=0.9,level distance=1.2cm,
			level 1/.style={sibling distance=4.5cm},
			level 2/.style={sibling distance=0.9cm}]
			
			\node[fill=black, shape = circle, draw, line width = 1pt, minimum size = 2.5mm, inner sep = 0mm] (root) at (0, 0){}
			child {node[fill=black, shape = circle, draw, line width = 1pt, minimum size = 2.5mm, inner sep = 0mm] (c1) {}
				child {node[shape = circle, draw, line width = 1pt, minimum size = 2.5mm, inner sep = 0mm] (c11) {}}
				child {node[shape = circle, draw, line width = 1pt, minimum size = 2.5mm, inner sep = 0mm] (c12) {}}
				child {node[shape = circle, draw, line width = 1pt, minimum size = 2.5mm, inner sep = 0mm] (c13) {}}
				child {node[shape = circle, draw, line width = 1pt, minimum size = 2.5mm, inner sep = 0mm] (c14) {}}
			}
			child {node[fill=black, shape = circle, draw, line width = 1pt, minimum size = 2.5mm, inner sep = 0mm] (c2) {}
				child {node[shape = circle, draw, line width = 1pt, minimum size = 2.5mm, inner sep = 0mm] (c21) {}}
				child {node[shape = circle, draw, line width = 1pt, minimum size = 2.5mm, inner sep = 0mm] (c22) {}}
				child {node[shape = circle, draw, line width = 1pt, minimum size = 2.5mm, inner sep = 0mm] (c23) {}}
				child {node[shape = circle, draw, line width = 1pt, minimum size = 2.5mm, inner sep = 0mm] (c24) {}}
			}
			child {node[fill=black, shape = circle, draw, line width = 1pt, minimum size = 2.5mm, inner sep = 0mm] (c3) {}
				child {node[shape = circle, draw, line width = 1pt, minimum size = 2.5mm, inner sep = 0mm] (c31) {}}
				child {node[shape = circle, draw, line width = 1pt, minimum size = 2.5mm, inner sep = 0mm] (c32) {}}
				child {node[shape = circle, draw, line width = 1pt, minimum size = 2.5mm, inner sep = 0mm] (c33) {}}
				child {node[shape = circle, draw, line width = 1pt, minimum size = 2.5mm, inner sep = 0mm] (c34) {}}
			}
			child {node[fill=black, shape = circle, draw, line width = 1pt, minimum size = 2.5mm, inner sep = 0mm] (c4) {}
				child {node[shape = circle, draw, line width = 1pt, minimum size = 2.5mm, inner sep = 0mm] (c41) {}}
				child {node[shape = circle, draw, line width = 1pt, minimum size = 2.5mm, inner sep = 0mm] (c42) {}}
				child {node[shape = circle, draw, line width = 1pt, minimum size = 2.5mm, inner sep = 0mm] (c43) {}}
				child {node[shape = circle, draw, line width = 1pt, minimum size = 2.5mm, inner sep = 0mm] (c44) {}}
			};
		
		\node (x1) at ($(c11.south) + (0,-0.4)$) {\huge\(x_1\)\normalsize};
		\node (x2) at ($(c12.south) + (0,-0.4)$) {\huge\(x_2\)\normalsize};
		\node (x3) at ($(c13.south) + (0,-0.4)$) {\huge\(x_3\)\normalsize};
		\node (x4) at ($(c14.south) + (0,-0.4)$) {\huge\(x_4\)\normalsize};
		
		\node (x5) at ($(c21.south) + (0,-0.4)$) {\huge\(x_5\)\normalsize};
		\node (x6) at ($(c22.south) + (0,-0.4)$) {\huge\(x_6\)\normalsize};
		\node (x7) at ($(c23.south) + (0,-0.4)$) {\huge\(x_7\)\normalsize};
		\node (x8) at ($(c24.south) + (0,-0.4)$) {\huge\(x_8\)\normalsize};
		
		\node (x9) at ($(c31.south) + (0,-0.4)$) {\huge\(x_9\)\normalsize};
		\node (x10) at ($(c32.south) + (0,-0.4)$) {\huge\(x_{10}\)\normalsize};
		\node (x11) at ($(c33.south) + (0,-0.4)$) {\huge\(x_{11}\)\normalsize};
		\node (x12) at ($(c34.south) + (0,-0.4)$) {\huge\(x_{12}\)\normalsize};
		
		\node (x13) at ($(c41.south) + (0,-0.4)$) {\huge\(x_{13}\)\normalsize};
		\node (x14) at ($(c42.south) + (0,-0.4)$) {\huge\(x_{14}\)\normalsize};
		\node (x15) at ($(c43.south) + (0,-0.4)$) {\huge\(x_{15}\)\normalsize};
		\node (x15) at ($(c44.south) + (0,-0.4)$) {\huge\(x_{16}\)\normalsize};
		
		\node (treeLabel) at ($(root.north) + (0.4,0.4)$) {\huge\(\T_\W\)\normalsize};
		\end{tikzpicture}
		\hfil
		\begin{tikzpicture}[scale=.9,every node/.style={minimum size=1cm}]
			
			\draw[step=1cm, black] (0,0) grid (4,4); 
			
			\node (x1) at (0.5,0.5) {\huge\(x_1\)\normalsize};
			\node (x2) at (1.5,0.5) {\huge\(x_2\)\normalsize};
			\node (x3) at (0.5,1.5) {\huge\(x_3\)\normalsize};
			\node (x4) at (1.5,1.5) {\huge\(x_4\)\normalsize};

			\node (x5) at (2.5,0.5) {\huge\(x_5\)\normalsize};
			\node (x6) at (3.5,0.5) {\huge\(x_6\)\normalsize};
			\node (x7) at (2.5,1.5) {\huge\(x_7\)\normalsize};
			\node (x8) at (3.5,1.5) {\huge\(x_8\)\normalsize};		
			
			\node (x9) at (0.5,2.5) {\huge\(x_9\)\normalsize};
			\node (x10) at (1.5,2.5) {\huge\(x_{10}\)\normalsize};
			\node (x11) at (0.5,3.5) {\huge\(x_{11}\)\normalsize};
			\node (x12) at (1.5,3.5) {\huge\(x_{12}\)\normalsize};	
			
			\node (x13) at (2.5,2.5) {\huge\(x_{13}\)\normalsize};
			\node (x14) at (3.5,2.5) {\huge\(x_{14}\)\normalsize};
			\node (x15) at (2.5,3.5) {\huge\(x_{15}\)\normalsize};
			\node (x16) at (3.5,3.5) {\huge\(x_{16}\)\normalsize};		
		\end{tikzpicture}
	\end{adjustbox}
	
	\vspace{10pt}
	
	\begin{adjustbox}{max size={0.46\textwidth}}
		\begin{tikzpicture}[scale=0.9, level distance=1.2cm,
			level 1/.style={sibling distance=4.5cm},
			level 2/.style={sibling distance=0.9cm}]
			
			\node[fill=black, shape = circle, draw, line width = 1pt, minimum size = 2.5mm, inner sep = 0mm] (root) at (0, 0){}
			child {node[fill=black, shape = circle, draw, line width = 1pt, minimum size = 2.5mm, inner sep = 0mm] (c1) {}
				child {node[shape = circle, draw, line width = 1pt, minimum size = 2.5mm, inner sep = 0mm] (c11) {}}
				child {node[shape = circle, draw, line width = 1pt, minimum size = 2.5mm, inner sep = 0mm] (c12) {}}
				child {node[shape = circle, draw, line width = 1pt, minimum size = 2.5mm, inner sep = 0mm] (c13) {}}
				child {node[shape = circle, draw, line width = 1pt, minimum size = 2.5mm, inner sep = 0mm] (c14) {}}
			}
			child {node[shape = circle, draw, line width = 1pt, minimum size = 2.5mm, inner sep = 0mm] (c2) {}
				child[black!15!] {node[shape = circle, draw, line width = 1pt, minimum size = 2.5mm, inner sep = 0mm] (c21) {}}
				child[black!15!] {node[shape = circle, draw, line width = 1pt, minimum size = 2.5mm, inner sep = 0mm] (c22) {}}
				child[black!15!] {node[shape = circle, draw, line width = 1pt, minimum size = 2.5mm, inner sep = 0mm] (c23) {}}
				child[black!15!] {node[shape = circle, draw, line width = 1pt, minimum size = 2.5mm, inner sep = 0mm] (c24) {}}
			}
			child {node[shape = circle, draw, line width = 1pt, minimum size = 2.5mm, inner sep = 0mm] (c3) {}
				child[black!15!] {node[shape = circle, draw, line width = 1pt, minimum size = 2.5mm, inner sep = 0mm] (c31) {}}
				child[black!15!] {node[shape = circle, draw, line width = 1pt, minimum size = 2.5mm, inner sep = 0mm] (c32) {}}
				child[black!15!] {node[shape = circle, draw, line width = 1pt, minimum size = 2.5mm, inner sep = 0mm] (c33) {}}
				child[black!15!] {node[shape = circle, draw, line width = 1pt, minimum size = 2.5mm, inner sep = 0mm] (c34) {}}
			}
			child {node[fill=black, shape = circle, draw, line width = 1pt, minimum size = 2.5mm, inner sep = 0mm] (c4) {}
				child {node[shape = circle, draw, line width = 1pt, minimum size = 2.5mm, inner sep = 0mm] (c41) {}}
				child {node[shape = circle, draw, line width = 1pt, minimum size = 2.5mm, inner sep = 0mm] (c42) {}}
				child {node[shape = circle, draw, line width = 1pt, minimum size = 2.5mm, inner sep = 0mm] (c43) {}}
				child {node[shape = circle, draw, line width = 1pt, minimum size = 2.5mm, inner sep = 0mm] (c44) {}}
			};
			
			\node (t1) at ($(c11.south) + (0,-0.4)$) {\huge\(t_1\)\normalsize};
			\node (t2) at ($(c12.south) + (0,-0.4)$) {\huge\(t_2\)\normalsize};
			\node (t3) at ($(c13.south) + (0,-0.4)$) {\huge\(t_3\)\normalsize};
			\node (t4) at ($(c14.south) + (0,-0.4)$) {\huge\(t_4\)\normalsize};
			
			\node[black!15!] (x5) at ($(c21.south) + (0,-0.4)$) {\huge\(x_5\)\normalsize};
			\node[black!15!] (x6) at ($(c22.south) + (0,-0.4)$) {\huge\(x_6\)\normalsize};
			\node[black!15!] (x7) at ($(c23.south) + (0,-0.4)$) {\huge\(x_7\)\normalsize};
			\node[black!15!] (x8) at ($(c24.south) + (0,-0.4)$) {\huge\(x_8\)\normalsize};
			
			\node[black!15!] (x9) at ($(c31.south) + (0,-0.4)$) {\huge\(x_9\)\normalsize};
			\node[black!15!] (x10) at ($(c32.south) + (0,-0.4)$) {\huge\(x_{10}\)\normalsize};
			\node[black!15!] (x11) at ($(c33.south) + (0,-0.4)$) {\huge\(x_{11}\)\normalsize};
			\node[black!15!] (x12) at ($(c34.south) + (0,-0.4)$) {\huge\(x_{12}\)\normalsize};
			
			\node (t7) at ($(c41.south) + (0,-0.4)$) {\huge\(t_{7}\)\normalsize};
			\node (t8) at ($(c42.south) + (0,-0.4)$) {\huge\(t_{8}\)\normalsize};
			\node (t9) at ($(c43.south) + (0,-0.4)$) {\huge\(t_{9}\)\normalsize};
			\node (t10) at ($(c44.south) + (0,-0.4)$) {\huge\(t_{10}\)\normalsize};
			
			\node (t5) at ($(c2.west) + (-0.4,0)$) {\huge\(t_{5}\)\normalsize};
			\node (t6) at ($(c3.east) + (0.4,0)$) {\huge\(t_{6}\)\normalsize};
			
			\node (treeLabel) at ($(root.north) + (0.4,0.4)$) {\huge\(\T\)\normalsize};
		\end{tikzpicture}
		\hfil
		\begin{tikzpicture}[scale=.9,every node/.style={minimum size=1cm}]
		
		\draw[step=1cm, black] (0,0) grid (2,2); 
		\draw[step=1cm, black] (2,2) grid (4,4); 
		\draw[step=2cm, black] (2,0) grid (4,2); 
		\draw[step=2cm, black] (0,2) grid (2,4); 
		
		\node (t1) at (0.5,0.5) {\huge\(t_1\)\normalsize};
		\node (t2) at (1.5,0.5) {\huge\(t_2\)\normalsize};
		\node (t3) at (0.5,1.5) {\huge\(t_3\)\normalsize};
		\node (t4) at (1.5,1.5) {\huge\(t_4\)\normalsize};
		
		\node (t7) at (2.5,2.5) {\huge\(t_{7}\)\normalsize};
		\node (t8) at (3.5,2.5) {\huge\(t_{8}\)\normalsize};
		\node (t8) at (2.5,3.5) {\huge\(t_{9}\)\normalsize};
		\node (t10) at (3.5,3.5) {\huge\(t_{10}\)\normalsize};	
		
		\node (t5) at (3,1) {\huge\(t_{5}\)\normalsize};	
		\node (t6) at (1,3) {\huge\(t_{6}\)\normalsize};	
	\end{tikzpicture}
	\end{adjustbox}
	
	\caption{Visual depiction of the tree \(\T_\W \in \T^\Q\) along with a tree \(\T \in \T^\Q\) and corresponding grid representations. Elements of the set \(\Nint(\cdot)\) are shown in black, whereas elements of the set \(\Nlf(\cdot)\) are white and numbered. Notice that the tree \(\T\) aggregates nodes \(x_5,~x_6,~x_7,~\text{and}~ x_8\) to \(t_5\) and the nodes \(x_5,~x_6,~x_7,~\text{and}~ x_8\) to \(t_6\). The nodes \(x_1,~x_2,~x_3,~\text{and}~x_4\) as well as \(x_{13},~x_{14},~x_{15},~\text{and}~x_{16}\) are not aggregated in this example. Aggregated nodes of \(\T_\W\) are shown in gray.}
	\label{fig:encoder_tree_equivalence}
\end{figure}
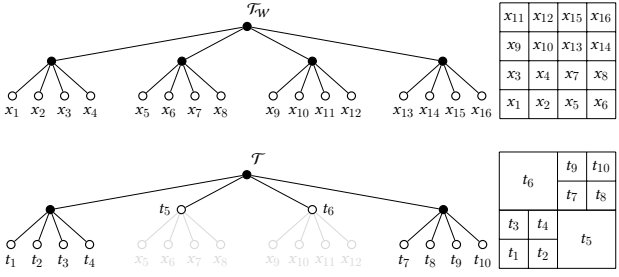

Let the set of real numbers be denoted \(\Re\), and, for any strictly positive integer \(n\), let \(\Re^n\) denote the \(n\)-dimensional Euclidean space.
For any vector \(\mathbf x \in \Re^n\), \([\mathbf x]_i\), \(i \in \{1,\ldots,n\}\), denotes the \(i^{\text{th}}\) element of \(\mathbf x\).
In this paper, we will restrict our attention to environment abstractions in the form of multi-resolution (e.g., pruned) quadtrees.
However, the contributions of this paper are valid for other, more general, tree structures.
The utility of multi-resolution quadtrees and hierarchical structures to represent environments is evidenced by their ubiquitous employment in robotics and autonomous systems \cite{Hauer2015,Hornung2013,Einhorn2011,Kambhampati1986}.
We assume that there exists an integer \(\ell > 0\) such that the world \(\W \subset \Re^2\) (generalizable to \(\Re^n\)), is contained within a square (hypercube) of side length \(2^{\ell}\). 
The environment \(\W\) is assumed to be a two-dimensional grid-world where each element of \(\W\) is a square (hypercube) of unit side length.
Furthermore, the quadtree corresponding to the finest resolution of \(\W\) is denoted \(\T_\W\); an example is shown in Figure \ref{fig:encoder_tree_equivalence}.
We will adopt the notation and conventions from \cite{Larsson2020}.

In this paper, we aim to generate information-theoretic graph abstractions, which requires the specification of a probability space as well as the introduction of fundamental information-theoretic quantities.
To this end, let \((\Omega, \mathcal F, \mathbb{P})\) be a probability space with finite sample space \(\Omega\), \(\sigma\)-algebra \(\mathcal F\) and probability measure \(\mathbb{P}\).
We define random variables \(X: \Omega \to \Re\), \(T: \Omega \to \Re\) and \(Y: \Omega \to \Re\), and take \(\Omega_X = \{x \in \Re : X(\omega) = x, \omega \in \Omega\}\) with \(\Omega_T\) and \(\Omega_Y\) defined analogously.
The probability mass function for the random variable \(X\) is denoted \(p(x) = \mathbb{P}(\{\omega \in \Omega : X(\omega) = x\})\), with the mass functions for \(T\) and \(Y\) defined similarly.
The Kullback-Liebler (KL) divergence between two distributions \(p(x)\) and \(\nu(x)\) is defined as 
\begin{equation}
	\mathrm{D}_{\mathrm{KL}}(p(x),\nu(x)) = \sum_x p(x) \log \frac{p(x)}{\nu(x)}.
\end{equation}
The mutual information between two random variables \(X\) and \(T\) is defined in terms of the KL-divergence as
\begin{equation} \label{eq:defnMI}
	I(T;X) = \mathrm{D}_{\mathrm{KL}}(p(t,x),p(t)p(x)).
\end{equation}
The mutual information captures the statistical dependence between the random variables \(X\) and \(T\) in that \(I(T;X) = 0\) if and only if \(p(t,x) = p(t)p(x)\)\footnote{This implies statistical independence of the two random variables \(X\) and \(T\).}. 
The mutual information can also be written as
\begin{equation} \label{eq:entropyExpansionMI}
	I(T;X) = H(T) - H(T|X) = H(X) - H(X|T),
\end{equation}
where \(H(T)\) and \(H(T|X)\) are the Shannon and conditional entropy, respectively\footnote{The Shannon and conditional entropy are given by \(H(T) = -\sum_t p(t)\log p(t)\) and \(H(T|X) = -\sum_{t,x} p(t,x) \log p(t|x)\), respectively.}.
The mutual information plays a vital role in information-theoretic frameworks for data compression.

We consider the Information Bottleneck (IB) method for data compression \cite{Tishby1999}.
The goal of the IB method is to form a compressed representation \(T\) of an original signal \(X\) while retaining as much information as possible regarding a relevant variable \(Y\) \cite{Tishby1999,GiladBachrach2003}.
To accomplish this objective, the IB framework formulates and optimization problem where the aim is to maximize the model quality, captured by the maximization of \(I(T;Y)\), subject to constraints on the degree of compression, quantified by \(I(T;X)\).
The IB method assumes \(p(t,x,y) = p(t|x)p(x,y)\), which is equivalent to assuming the joint distribution \(p(t,x,y)\) factors according to the Markov Chain \(T \leftrightarrow X \leftrightarrow Y\) and implies \(I(T;Y) \leq I(X;Y)\) according to the data processing inequality \cite{Cover2006}.
The IB problem seeks an encoder \(p(t|x)\) so as to solve the problem
\begin{equation} \label{eq:IBobj_1}
	\max_{p(t|x)} I(T;Y) \hspace{8pt} \text{s.t.} \hspace{8pt} I(T;X) \leq D,
\end{equation}
where the minimization is over all conditional distributions \(p(t|x)\) and \(D \geq 0\) is a given constant \cite{GiladBachrach2003}.
The problem \eqref{eq:IBobj_1} has Lagrangian
\begin{equation} \label{eq:originalIB_Lagrangian}
	\max_{p(t|x)} I(T;Y) - \frac{1}{\beta} I(T;X),
\end{equation}
where \(\beta > 0\) is a parameter that balances the importance of compression and relevant information retention when forming the compressed representation \(T\) \cite{Tishby1999}.
An implicit solution to \eqref{eq:originalIB_Lagrangian} can be analytically derived as a function of \(\beta\), and an iterative algorithm can be employed to obtain locally optimal solutions for given values of \(\beta\) \cite{Tishby1999}.
As \(\beta\) is varied, the solution to \eqref{eq:originalIB_Lagrangian} trades the importance of information retention (maximization of \(I(T;Y)\)) and compression (minimization of \(I(T;X)\)).
However, not all encoders \(p(t|x)\) correspond to  valid tree representations, as the tree structure places constraints on the encoder \(p(t|x)\), presenting a significant challenge.

The connection between optimal encoder problems, the IB method and hierarchical tree structures was recently explored in \cite{Larsson2020}.
However, in contrast to the work presented in this paper, the authors of \cite{Larsson2020} solve the IB problem over the space of multi-resolution trees as a function of \(\beta\), and do not discuss a method for obtaining \(\beta\) when given a value of \(D\).
We argue that for agents with resource constraints, it is more practical to specify a value of \(D\), which is directly representative of the agents resource limitations and is independent of the input distribution \(p(x,y)\), than it is to specify a value of \(\beta\), which may give rise to different amounts of relevant information \(I(T;Y)\) (i.e., different representations) by simply varying \(p(x,y)\).

Interestingly, it was observed in \cite{Larsson2020} that each tree \(\T_q \in \T^\Q \) corresponds to a deterministic encoder \(p_q(t|x)\) that maps leafs \(x \in \Nlf(\T_\W)\), to the leafs \(t \in \Nlf(\T_q)\).
That is, \(\T_q \in \T^\Q\) specifies an encoder \(p_q(t|x)\) where, for all \(t \in \Nlf(\T_q)\) and \(x \in \Nlf(\T_\W)\), \(p_q(t|x) \in \{0,1\}\) and \(p_q(t|x) = 1\) only if the node \(x \in \Nlf(\T_\W)\) is aggregated to the node \(t \in \Nlf(\T_q)\).
An example is presented in Figure \ref{fig:encoder_tree_equivalence}.
Thus, by selecting different trees \(\T_q \in \T^\Q\), we alter the set \(\Nlf(\T_q)\), thereby changing the multi-resolution representation of the world \(\W\). 
Our goal is to use the IB method to select the multi-resolution tree that is most informative regarding the relevant variable \(Y\).

Given any tree \(\T_q \in \T^\Q\), the resulting compressed representation, denoted \(T_q:\Omega \to \Nlf(\T_q)\), is known and the joint distribution is determined according to \(p_q(t,x,y) = p_q(t|x)p(x,y)\).
As a result, the values of \(I(T_q;X)\) and \(I(T_q;Y)\) can be computed.
We then define the functions \(I_X : \T^\Q \to [0,\infty)\) and \(I_Y: \T^\Q \to [0,\infty)\) as \(I_X(\T_q) = I(T_q;X)\) and \(I_Y(\T_q) = I(T_q;Y)\), respectively.
We now formally state our problem. 
\begin{problem} \label{prob:IBtreeOptimProblem1}
	Given the joint distribution \(p(x,y)\), a scalar \(D \geq 0\), and the world \(\W\), we consider the problem
	\begin{equation} \label{eq:IBtreeOptim1}
		\max_{\T_q \in \T^\Q} I_Y(\T_q),
	\end{equation}
	subject to the constraint
	\begin{equation} \label{eq:IBtreeCons1}
		I_X(\T_q) \leq D.
	\end{equation}
\end{problem}
To the best of our knowledge, no method currently exists for solving Problem \ref{prob:IBtreeOptimProblem1} for a given value of \(D\).
The most closely related work is that of \cite{Larsson2020}, where the authors develop a framework for multi-resolution tree abstractions by solving the problem \eqref{eq:originalIB_Lagrangian} as a function of \(\beta\). 
In doing so, the authors of \cite{Larsson2020} utilize a recursive function, akin to Q-functions in reinforcement learning.
In contrast, our contribution in this paper is to develop a method for solving the problem \eqref{eq:IBtreeOptim1} subject to \eqref{eq:IBtreeCons1} as a function of \(D\geq 0\) directly, without the need to employ recursive functions or specify trade-off parameters.

\section{Solution Approach}
The constrained optimization problem defined by \eqref{eq:IBtreeOptim1} and \eqref{eq:IBtreeCons1} can be solved by an exhaustive search over the space of multi-resolution tree abstractions.
However, for larger environments, the application of an exhaustive search is intractable as the space of feasible multi-resolution trees is vast.
Therefore, we aim to develop an alternative approach that does not require the generation of all candidate solutions when searching for a solution to Problem \ref{prob:IBtreeOptimProblem1}.

\begin{figure}
	\centering
	\begin{adjustbox}{max size={0.37\textwidth}}
		\begin{tikzpicture}[scale=1,level distance=1.2cm,
			level 1/.style={sibling distance=4.5cm},
			level 2/.style={sibling distance=0.9cm}]	
			\node[shape = circle, draw, line width = 1pt, minimum size = 2.5mm, inner sep = 0mm] (root1) at (0,0) {};        
			
			\node[fill=black, shape = circle, draw, line width = 1pt, minimum size = 2.5mm, inner sep = 0mm] (root2) at (0,-2) {}
			child {node[shape = circle, draw, line width = 1pt, minimum size = 2.5mm, inner sep = 0mm] {}
			}
			child {node[shape = circle, draw, line width = 1pt, minimum size = 2.5mm, inner sep = 0mm] {}
			}
			child {node[shape = circle, draw, line width = 1pt, minimum size = 2.5mm, inner sep = 0mm] {}
			}
			child {node[shape = circle, draw, line width = 1pt, minimum size = 2.5mm, inner sep = 0mm] {}
			};	
			\node[fill=black, shape = circle, draw, line width = 1pt, minimum size = 2.5mm, inner sep = 0mm] (root3) at (0,-4) {}
			child {node[fill=black, shape = circle, draw, line width = 1pt, minimum size = 2.5mm, inner sep = 0mm] {}
				child {node[shape = circle, draw, line width = 1pt, minimum size = 2.5mm, inner sep = 0mm] {}}
				child {node[shape = circle, draw, line width = 1pt, minimum size = 2.5mm, inner sep = 0mm] {}}
				child {node[shape = circle, draw, line width = 1pt, minimum size = 2.5mm, inner sep = 0mm] {}}
				child {node[shape = circle, draw, line width = 1pt, minimum size = 2.5mm, inner sep = 0mm] {}}
			}
			child {node[shape = circle, draw, line width = 1pt, minimum size = 2.5mm, inner sep = 0mm] {}
			}
			child {node[shape = circle, draw, line width = 1pt, minimum size = 2.5mm, inner sep = 0mm] {}
			}
			child {node[shape = circle, draw, line width = 1pt, minimum size = 2.5mm, inner sep = 0mm] {}
			};
			\node[fill=black, shape = circle, draw, line width = 1pt, minimum size = 2.5mm, inner sep = 0mm] (root4) at (0, -6){}
			child {node[fill=black, shape = circle, draw, line width = 1pt, minimum size = 2.5mm, inner sep = 0mm] {}
				child {node[shape = circle, draw, line width = 1pt, minimum size = 2.5mm, inner sep = 0mm] {}}
				child {node[shape = circle, draw, line width = 1pt, minimum size = 2.5mm, inner sep = 0mm] {}}
				child {node[shape = circle, draw, line width = 1pt, minimum size = 2.5mm, inner sep = 0mm] {}}
				child {node[shape = circle, draw, line width = 1pt, minimum size = 2.5mm, inner sep = 0mm] {}}
			}
			child {node[shape = circle, draw, line width = 1pt, minimum size = 2.5mm, inner sep = 0mm] {}
			}
			child {node[shape = circle, draw, line width = 1pt, minimum size = 2.5mm, inner sep = 0mm] {}
			}
			child {node[fill=black, shape = circle, draw, line width = 1pt, minimum size = 2.5mm, inner sep = 0mm] {}
				child {node[shape = circle, draw, line width = 1pt, minimum size = 2.5mm, inner sep = 0mm] {}}
				child {node[shape = circle, draw, line width = 1pt, minimum size = 2.5mm, inner sep = 0mm] {}}
				child {node[shape = circle, draw, line width = 1pt, minimum size = 2.5mm, inner sep = 0mm] {}}
				child {node[shape = circle, draw, line width = 1pt, minimum size = 2.5mm, inner sep = 0mm] {}}
			};
			\draw [->, >=stealth, line width=0.5mm, black] ($(root1.south) + (0,-0.2)$) -- ($(root2.north) + (0,0.2)$);
			
			\draw [->, >=stealth, line width=0.5mm, black] ($(root2.south) + (0,-0.2)$) -- ($(root3.north) + (0,0.2)$);

			\draw [->, >=stealth, line width=0.5mm, black] ($(root3.south) + (0,-0.2)$) -- ($(root4.north) + (0,0.2)$);
			\node (tree1Label) at ($(root1) + (9,0)$) {\Huge \(\mathcal T_0\) \normalsize}; 
			\node (tree2Label) at ($(root2) + (9,0)$) {\Huge \(\mathcal T_1\) \normalsize}; 
			\node (tree3Label) at ($(root3) + (9,0)$) {\Huge \(\mathcal T_2\) \normalsize}; 
			\node (tree4Label) at ($(root4) + (9,0)$) {\Huge \(\mathcal T_3\) \normalsize}; 
			
			\draw [->, >=stealth, line width=0.5mm, black] ($(tree1Label.south) + (0,-0.2)$) -- ($(tree2Label.north) + (0,0.2)$); 
			\draw [->, >=stealth, line width=0.5mm, black] ($(tree2Label.south) + (0,-0.2)$) -- ($(tree3Label.north) + (0,0.2)$); 
			\draw [->, >=stealth, line width=0.5mm, black] ($(tree3Label.south) + (0,-0.2)$) -- ($(tree4Label.north) + (0,0.2)$); 
		\end{tikzpicture}
	\end{adjustbox}
	\caption{Sequence of trees \(\left\{ \mathcal T_i \right\}_{i=0}^{3} \subseteq \mathcal T^{\mathcal Q}\) leading from \(\mathcal T_0 = \mathcal R_\W\), the root of \(\T_\W\), to the tree \(\mathcal T_3\).
	Note that \(\mathcal N\left( \mathcal T_{i+1}\right) \setminus \mathcal N\left( \mathcal T_{i}\right) = \mathcal C(t)\) for some \(t \in \Nlf\left( \mathcal T_i \right)\) holds all \(i \in \{0,1,2\}\). For each tree \(\T_i\), \(i \in \{0,1,2,3\}\), the set of interior nodes, \(\Nint(\T_i)\), is colored black whereas nodes in \(\Nlf(\T_i)\) are white. For the tree \(\mathcal{R}_\W\), the set of interior nodes is empty.}
	\label{fig:sequenceOfTrees}
\end{figure}
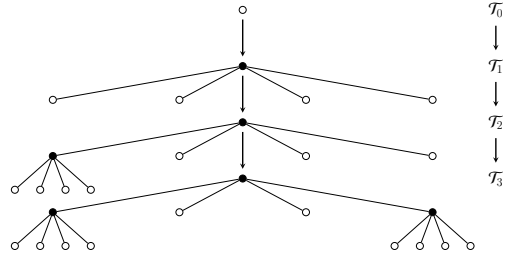

To this end, we note that for any tree \(\T_q \in \T\), the quantities \(I_X(\T_q)\) and \(I_Y(\T_q)\) can be written as
\begin{equation} \label{eq:seqenIxTree}
	I_X(\T_q) = I_X(\T_0) + \sum_{i=0}^{q-1} \left[ I_X(\T_{i+1}) - I_X(\T_i) \right],
\end{equation}
and
\begin{equation} \label{eq:seqenIyTree}
	I_Y(\T_q) = I_Y(\T_0) + \sum_{i=0}^{q-1} \left[ I_Y(\T_{i+1}) - I_Y(\T_i) \right],
\end{equation}
respectively.
Relations \eqref{eq:seqenIxTree} and \eqref{eq:seqenIyTree} allow us to express the value of \(I_X(\T_q)\) and \(I_Y(\T_q)\) in terms of a sequence of trees \(\{\T_0,\ldots,\T_{q-1},\T_q\}\), where \(\T_i \in \T^{\Q}\) for all \(i \in \{0,\ldots,q\}\).
Strictly speaking, the validity of \eqref{eq:seqenIxTree} and \eqref{eq:seqenIyTree} does not depend on the sequence \(\{\T_i\}_{i=0}^{q}\).
However, by selecting the sequence \(\{\T_i\}_{i=0}^{q}\) in a specific way, we obtain tractable expressions for \(I_X(\T_q)\) and \(I_Y(\T_q)\), as follows.

Note that any tree \(\T_q \in \T^\Q\) can be obtained by starting at the root of \(\T_\W\), denoted \(\mathcal R_\W \in \T^\Q\), and performing a sequence of nodal expansions.
More precisely, any \(\T_q \in \T^\Q\) can be obtained by a sequence \(\{\T_i\}_{i=0}^{q} \subseteq \T^\Q\), where \(\T_0 = \mathcal R_\W\) and \(\N(\T_{i+1}) \setminus \N(\T_i) = \chd(t)\) for some \(t \in \Nlf(\T_i)\) holds for all \(i \in \{0,\ldots,q-1\}\).
An illustrative example is shown in Figure \ref{fig:sequenceOfTrees}.
When considering sequences \(\{\T_i\}_{i=0}^{q}\) that satisfy \(\N(\T_{i+1}) \setminus \N(\T_i) = \chd(t)\) for some \(t \in \Nlf(\T_i)\) and all \(i \in \{0,\ldots,q-1\}\), we note that the difference between trees \(\T_{i+1}\) and  \(\T_i\) is the aggregation of the nodes \(\chd(t) \subseteq \Nlf(\T_{i+1})\) to their parent node \(t \in \Nlf(\T_i)\).
To quantify the change in information between the trees \(\T_{i+1}\) and \(\T_i\) in this case, we define \(\Delta I_X(\T_{i+1},\T_i)\) as
\begin{equation}
	\Delta I_X(\T_{i+1}, \T_i) = I_X(\T_{i+1}) - I_X(\T_{i}).
\end{equation}
Since the aggregation is deterministic, we have that \(p_q(t|x) \in \{0,1\}\) for all \(t \in \Nlf(\T_q)\), \(x \in \Nlf(\T_\W)\) and \(\T_q \in \T^\Q\), implying \(H(T_q|X) = 0\). 
Then, from \eqref{eq:entropyExpansionMI}, we obtain
\begin{equation} \label{eq:deltaIxTrees1}
	\Delta I_X(\T_{i+1}, \T_i) = H(T_{i+1}) - H(T_i).
\end{equation}
Through a direct calculation, it can be shown that \eqref{eq:deltaIxTrees1} is given by
\begin{equation} \label{eq:deltaIxTreeAndNodes1}
	\Delta I_X(\T_{i+1}, \T_i) = p(t) H(\Pi),
\end{equation}
where the node \(t \in \Nlf(\T_i)\) is expanded, adding \(\chd(t) \subseteq \Nlf(\T_{i+1})\) to the tree \(\T_i\) to create \(\T_{i+1}\), \(H(\Pi)\) is entropy of the distribution \(\Pi\)\footnote{By abuse of notation, for any discrete distribution \(\Pi=[\pi_1,\ldots,\pi_u]\), we write \(H(\Pi)\) to mean \(H(\Pi) = -\sum_{i=1}^{u}\pi_i \log \pi_i\).}, where \(\Pi \in \Re^{4}\) is given by
\begin{equation}
	\Pi = \left[\frac{p(t'_1)}{p(t)},\ldots, \frac{p(t'_{4})}{p(t)}\right],
\end{equation}
\(\chd(t) = \{t'_1,\ldots, t'_{4}\}\), and
\begin{equation}
	p(t) = \sum_{i=1}^{4} p(t'_i).
\end{equation}
In a similar spirit, we define \(\Delta I_Y(\T_{i+1}, \T_i)\) between the trees \(\T_{i+1}\) and \(\T_i\), where \(\N(\T_{i+1}) \setminus \N(\T_i) = \chd(t)\) for some \(t \in \Nlf(\T_i)\), as
\begin{equation} \label{eq:deltaIyTrees1}
	\Delta I_Y(\T_{i+1}, \T_i) = I_Y(\T_{i+1}) - I_Y(\T_i).
\end{equation}
One can show that \eqref{eq:deltaIyTrees1} is equivalent to
\begin{equation}\label{eq:deltaIyTreeAndNodes1}
	\Delta I_Y(\T_{i+1}, \T_{i}) = p(t) \mathrm{JS}_{\Pi}(p(y|t'_1),\ldots,p(y|t'_{4})),
\end{equation}
where \(\mathrm{JS}_\Pi(p(y|t'_1),\ldots,p(y|t'_{4}))\) is the Jensen-Shannon (JS) divergence \cite{Lin1991} defined as 
\begin{equation}
	\mathrm{JS}_{\Pi}(p(y|t'_1),\ldots,p(y|t'_{4})) = \sum_{i=1}^{4} [\Pi]_i \mathrm{D}_{\mathrm{KL}}(p(y|t'_i),p(y|t)),
\end{equation}
where 
\begin{equation}\label{eq:abstract_py_t}
	p(y|t) = \sum_{i=1}^{4} [\Pi]_i p(y|t'_i).
\end{equation}

It is important to note that the expressions \eqref{eq:deltaIxTreeAndNodes1} and \eqref{eq:deltaIyTreeAndNodes1} are functions only of the node expanded in moving from the tree \(\T_i\) to the tree \(\T_{i+1}\).
This observation is important for two reasons: (i) it allows for the value of \(\Delta I_X(\T_{i+1},\T_i)\) and \(\Delta I_Y(\T_{i+1},\T_i)\) to be computed without summing over the sample space of \(X\), which may be large, and (ii) the value of \(\Delta I_X(\T_{i+1},\T_i)\) and \(\Delta I_Y(\T_{i+1},\T_i)\) depends only on the node \(t \in \Nlf(\T_i)\) expanded to create \(\T_{i+1}\), and not the configuration of the other nodes in the tree \(\T_i\).
As a consequence of (ii), it follows that how one arrives at the tree \(\T_{i}\) is irrelevant.
Thus, we can write \(\Delta I_X(\T_{i+1},\T_i) = \Delta \hat{I}_X(t)\) and \(\Delta I_Y(\T_{i+1},\T_i) = \Delta \hat{I}_Y(t)\), where the node \(t \in \Nlf(\T_i)\) is expanded in moving from \(\T_i\) to \(\T_{i+1}\), and the functions \(\Delta \hat I_X: \Nint(\T_\W) \to [0,\infty)\) and \(\Delta \hat I_Y: \Nint(\T_\W) \to [0,\infty)\) are defined as 
\begin{equation} \label{eq:nodeWiseDeltaX}
	\Delta \hat I_X(t) = p(t) H(\Pi),
\end{equation}
and 
\begin{equation} \label{eq:nodeWiseDeltaY}
	\Delta \hat I_Y(t) = p(t) \mathrm{JS}_{\Pi}(p(y|t'_1),\ldots,p(y|t'_{4})),
\end{equation}
respectively.

What remains is to obtain expressions for \(I_X(\T_0)\) and \(I_Y(\T_0)\) in \eqref{eq:seqenIxTree} and \eqref{eq:seqenIyTree}.
To this end, notice that \(\mathcal R_\W\) corresponds to an encoder that aggregates all nodes \(x \in \Nlf(\T_\W)\) to the single node \(t = \Nlf(\mathcal R_\W)\).
As a result, the random variable \(T_0\) corresponding to the tree \(\T_0 = \mathcal{R}_\W\) has no entropy (\(H(T_0) = 0\)) since \(T_0\) has a single outcome with probability one.
Then, from \eqref{eq:entropyExpansionMI}, we see \(0 \leq I_X(\T_0) \leq H(T_0)\) and \(0 \leq I_Y(\T_0) \leq H(T_0)\), resulting with \(I_X(\T_0) = 0\) and \(I_Y(\T_0) = 0\). 
Lastly, notice that all sequences \(\{\T_i\}_{i=0}^{q}\) that lead to \(\T_q \in \T^\Q\) and satisfying \(\T_0 = \mathcal R_{\W}\) and \(\N(\T_{i+1})\setminus \N(\T_i) = \chd(t)\) for some \(t \in \Nlf(\T_i)\) and all \(i \in \{0,\ldots,q-1\}\), must expand all the interior nodes of \(\T_q\).
Combining the above observations with relations \eqref{eq:deltaIxTreeAndNodes1}, \eqref{eq:deltaIyTreeAndNodes1}, \eqref{eq:nodeWiseDeltaX} and \eqref{eq:nodeWiseDeltaY}, we obtain that \eqref{eq:seqenIxTree} and \eqref{eq:seqenIyTree} are given by
\begin{equation} \label{eq:treeXinformationFunction}
	I_X(\mathcal T_q) = \sum_{t \in \Nint(\T_q)} \Delta \hat I_X(t),
\end{equation}
and
\begin{equation} \label{eq:treeYinformationFunction}
	I_Y(\mathcal T_q) = \sum_{t \in \Nint(\T_q)} \Delta \hat I_Y(t).
\end{equation}

Relations \eqref{eq:treeXinformationFunction} and \eqref{eq:treeYinformationFunction} allow us to evaluate \(I_X(\T_q)\) and \(I_Y(\T_q)\) given the interior node set, \(\Nint(\T_q)\).
Furthermore, observe that the interior node set \(\Nint(\T_q)\) completely characterizes the tree \(\T_q \in \T^\Q\).
In other words, from only knowledge of  the set \(\Nint(\T_q)\), one can uniquely construct the tree \(\T_q \in\T^\Q\).
Thus, Problem \ref{prob:IBtreeOptimProblem1} can be reformulated as a search for interior nodes.

To formalize Problem \ref{prob:IBtreeOptimProblem1} as a search for interior nodes, we utilize the fact that any tree \(\T_q \in \T^\Q\) is specified by its interior node set.
With this in mind, notice that any tree \(\T_q \in \T^\Q\) can be represented by a vector \(\z \in \Re^{\lvert \Nint(\T_\W) \rvert}\), where the entries of \(\z\) indicate whether or not an expandable node \(t \in \Nint(\T_\W)\) is a member of the set \(\Nint(\T_q)\).
That is, for \(\T_q \in \T^\Q\) and \(t \in \Nint(\T_\W)\), \([\z]_t = 1\) implies \(t \in \Nint(\T_q)\), whereas \([\z]_t = 0\) indicates \(t \notin \Nint(\T_q)\).
Next, we define vectors \(\Delta_X,\Delta_Y \in \Re^{\lvert \Nint(\T_\W)\rvert}\) such that \([\Delta_X]_{t} = \Delta \hat I_X(t)\) and \([\Delta_Y]_{t} = \Delta \hat I_Y(t)\), and see from \eqref{eq:treeXinformationFunction} and \eqref{eq:treeYinformationFunction} that for any \(\T_q\in \T^\Q\), \(I_X(\T_q) = \z^{\tp} \Delta_X\) and \(I_Y(\T_q) = \z^{\tp} \Delta_Y\).
The IB problem over the space of multi-resolution trees is then formulated as the MILP
\begin{equation} \label{eq:intLinProg1}
	\max_{\z} ~ \z^{\tp} \Delta_Y,
\end{equation}
subject to the constraints
\begin{align}
	\z^{\tp} \Delta_X &\leq D, \label{eq:intLinProg1_Cons1} \\
	[\z]_{t'} - [\z]_t &\leq 0, ~ t \in \mathcal{B},~ t' \in \chd(t), \label{eq:intLinProg1_Cons2}\\
	[\z]_t &\in \{0,1\}, ~ t \in \Nint(\T_\W), \label{eq:intLinProg1_Cons3}
\end{align}
where \(D \geq 0\) and \(\mathcal{B} = \{ t \in \Nint(\T_\W) : \chd(t) \cap \Nint(\T_\W) \neq \varnothing \} \).
The constraint \eqref{eq:intLinProg1_Cons2} ensures that the vector \(\z\) correspond to a valid tree representation of \(\W\) by not allowing the children \(t' \in \chd(t)\) of some \(t \in \Nint(\T_\W)\) to be expanded unless their parent \(t\) is (i.e., if \([\z]_t = 0\) then \([\z]_{t'} = 0\)).
Notice that the above discussion pertains only to those nodes \(t \in \Nint(\T_\W)\) that have expandable children, which are those nodes in the set \(\mathcal B\).

The constraint \eqref{eq:intLinProg1_Cons1} has an information-theoretic interpretation in that it bounds the code rate of the compressed representation \cite{Cover2006,GiladBachrach2003}. 
Thus, our approach allows us to tailor the multi-resolution representation of \(\W\) according to the agent's on-board communication resources.

The problem \eqref{eq:intLinProg1}-\eqref{eq:intLinProg1_Cons3} can also be formulated as 
\begin{equation} \label{eq:intLinProg2}
    \min_{\z} ~ \z^{\tp} \Delta_X
\end{equation}
subject to 
\begin{align}
    \z^{\tp} \Delta_Y &\geq \hat D, \label{eq:intLinProg2_Cons1}\\
    [\z]_{t'} - [\z]_{t} &\leq 0, ~ t\in \mathcal{B},~ t' \in \chd(t), \label{eq:intLinProg2_Cons2}\\
    [\z]_t &\in \{0,1\}, ~t \in \Nint(\T_\W), \label{eq:intLinProg2_Cons3}
\end{align}
where \(\hat D \geq 0\).
The problem \eqref{eq:intLinProg2}-\eqref{eq:intLinProg2_Cons3} maximizes compression subject to a lower bound on the retained relevant information.
The formulations are equivalent and offer two perspectives on the same compression principle \cite{GiladBachrach2003}.
We emphasize that the utility of our approach is the ability to obtain tree abstractions that are explicitly constrained (by either \(D\) or \(\hat D\)) independently of \(p(x,y)\) by solving a linear program.
Although the optimization problem is a MILP that grows with order \(\mathcal O(\lvert \Nlf(\T_\W) \rvert)\) (the number of finest-resolution cells), one may consider convex relaxations of the problem by replacing the constraint \eqref{eq:intLinProg1_Cons3} with \([\z]_t \in [0,1]\) for \(t \in \Nint(\T_\W)\).
We will provide more detail regarding the use of convex relaxation to solve the integer-linear programs in the numerical example that follows.

\section{Numerical Example \& Discussion}

\begin{figure}[t]
	\centering
	\subfloat[]{\includegraphics[width=0.15\textwidth]{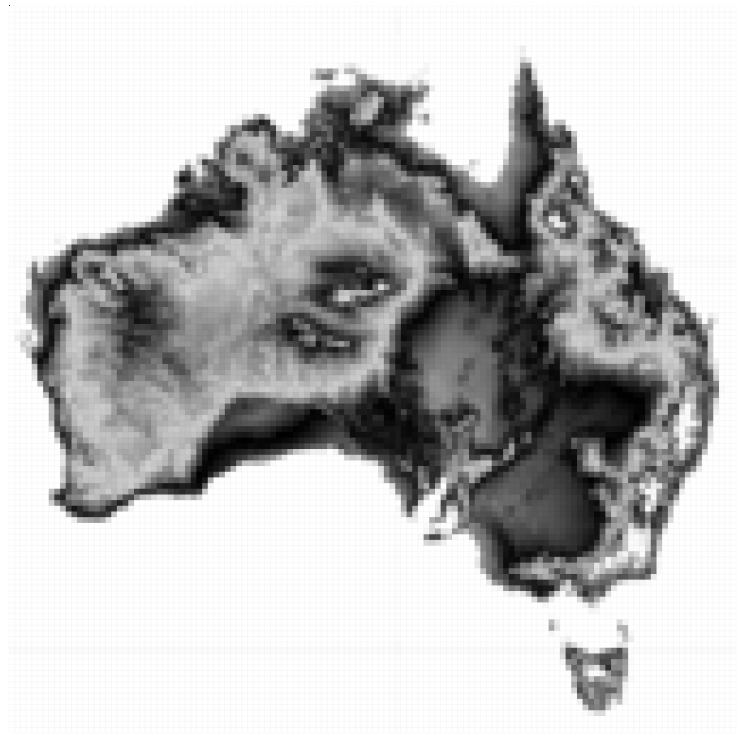}\label{fig:originalEnvW}}
	\hfil
	\subfloat[]{\includegraphics[width=0.15\textwidth]{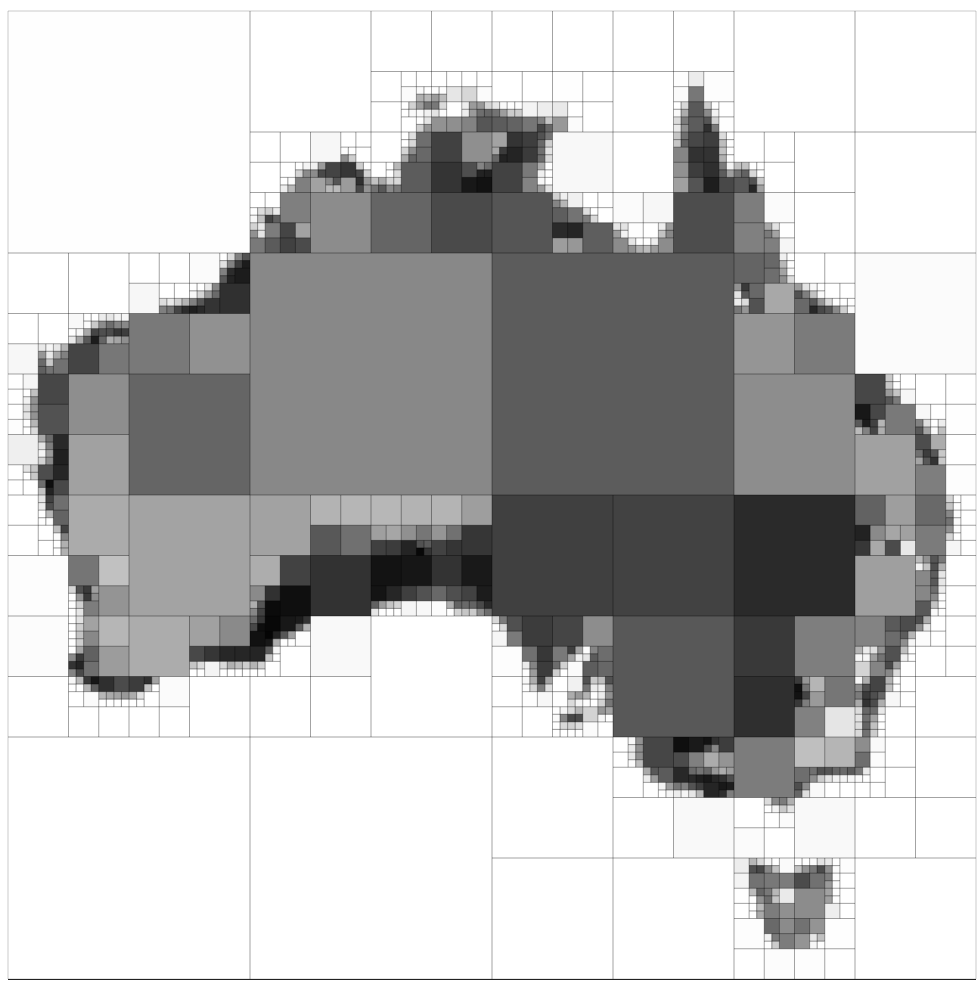}\label{fig:envabs1}}\\
	\subfloat[]{\includegraphics[width=0.15\textwidth]{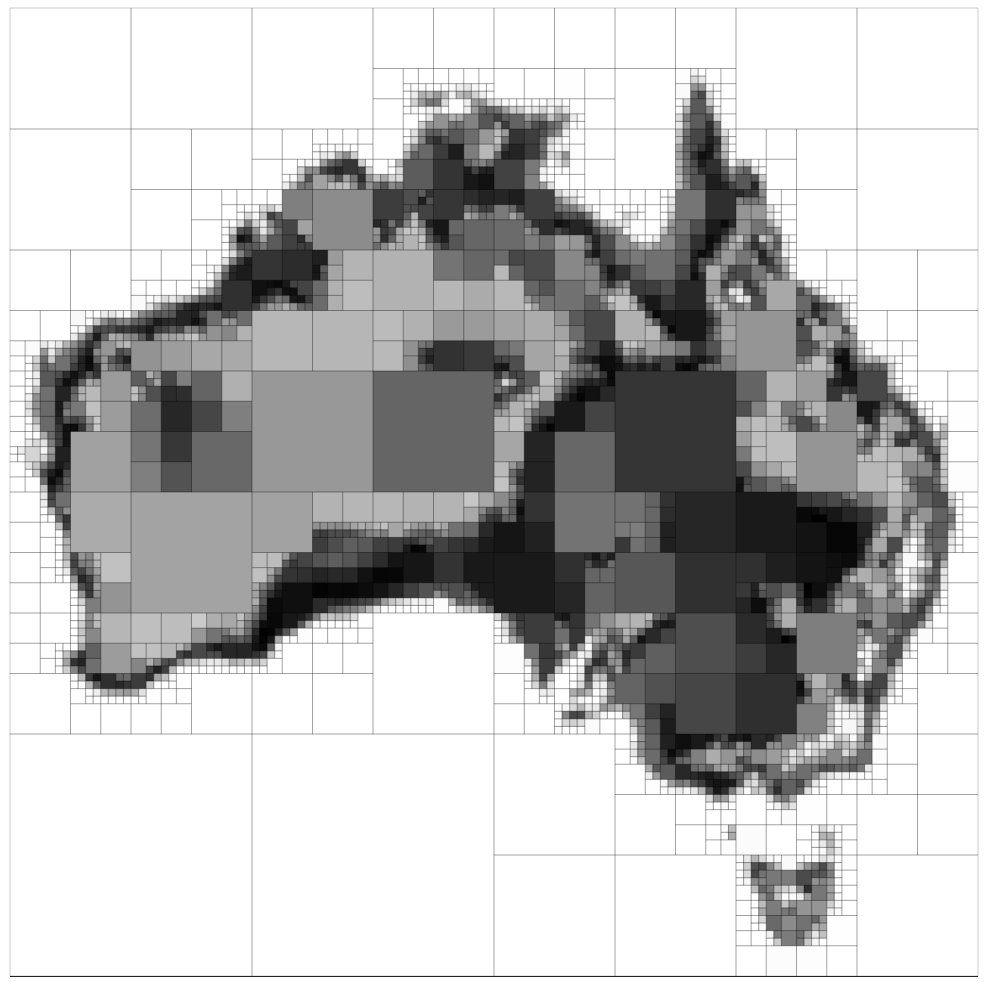}\label{fig:envabs2}}
	\hfil
	\subfloat[]{\includegraphics[width=0.15\textwidth]{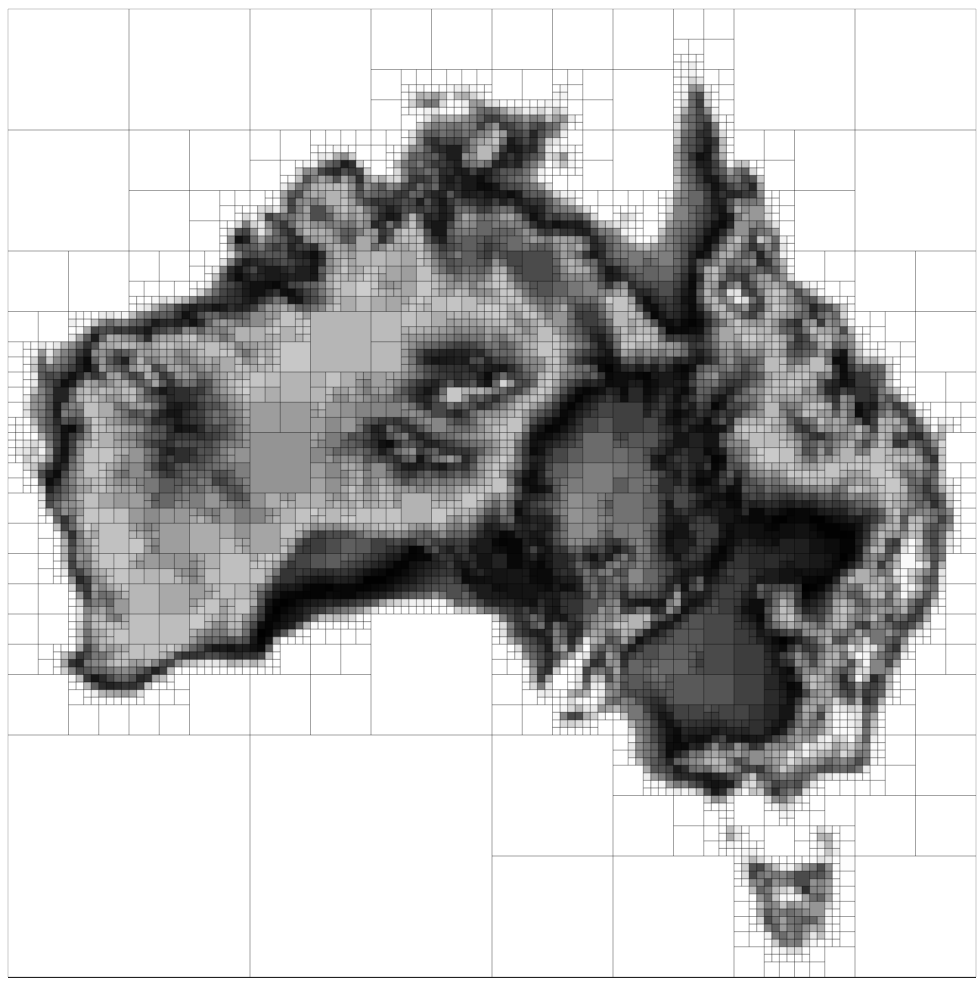}\label{fig:envabs3}}
	\caption{(a) original \(128 \times 128\) environment, (b)-(d) visualizations of abstractions obtained by solving the program \eqref{eq:intLinProg2}-\eqref{eq:intLinProg2_Cons3} with \(\hat D = 0.2745 ~\text{nats}\), \(\hat D = 0.2991 ~\text{nats}\), and \(\hat D = 0.3064 ~\text{nats}\), respectively. In this example, \(I(X;Y) = 0.3072 ~\text{nats}\), whereas \(I_Y(\T)\) values for the abstractions are: (b) \(0.2745 ~\text{nats}\) (\(7.6 \%\) of the original nodes), (c) \(0.2991 ~\text{nats}\) (\(18.5 \%\) of the original nodes), and (d) \(0.3064 ~\text{nats}\) (\(41.8 \%\) of the original nodes).}
\end{figure}

We consider generating multi-resolution abstractions for the \(128 \times 128\) (\(\ell = 7\)) satellite image shown in Figure \ref{fig:originalEnvW}.
The elevation information is assumed to be noisy, stemming from, for example, the agent's limited computational resources in processing camera images to create the map \cite{Wang2019}.
In this example, the relevant random variable \(Y:\Omega \to \{0,1\}\) represents the elevation encoded by the map, where \(Y = 1\) corresponds to higher elevations, and \(Y= 0\) is sea-level.
The color intensity of the map in Figure \ref{fig:originalEnvW} depicts the distribution \(p(y=1|x)\), where the individual cells are the outcomes of \(X\).
An agent may need to abstract the map due to its limited communication resources (as a function of \(D\)), or to limit the amount of information processing required to distinguish features in the environment (as a function of \(\hat D\)).

The Gurobi CVX solver \cite{cvx,cvx2} is used to obtain solutions to \eqref{eq:intLinProg2}-\eqref{eq:intLinProg2_Cons3}, resulting in the abstractions shown in Figures \ref{fig:envabs1}-\ref{fig:envabs3}.
As the value of \(\hat D\) is increased, we see from Figure \ref{fig:envabs1}-\ref{fig:envabs3} that the environment resolution increases and approaches that of the original map.
Observe that at lower values of \(\hat D\), our framework reveals the more salient details of the map; namely regions where there is a stark contrast in color intensity between neighboring cells. 
Furthermore, as \(\hat D\) is increased, more details of the map are revealed, whereas areas of constant \(p(y=1|x)\) remain aggregated (e.g., the lower left portion of the map in Figures \ref{fig:envabs1}-\ref{fig:envabs3}).

\begin{figure}[b]
	\includegraphics[scale=0.28]{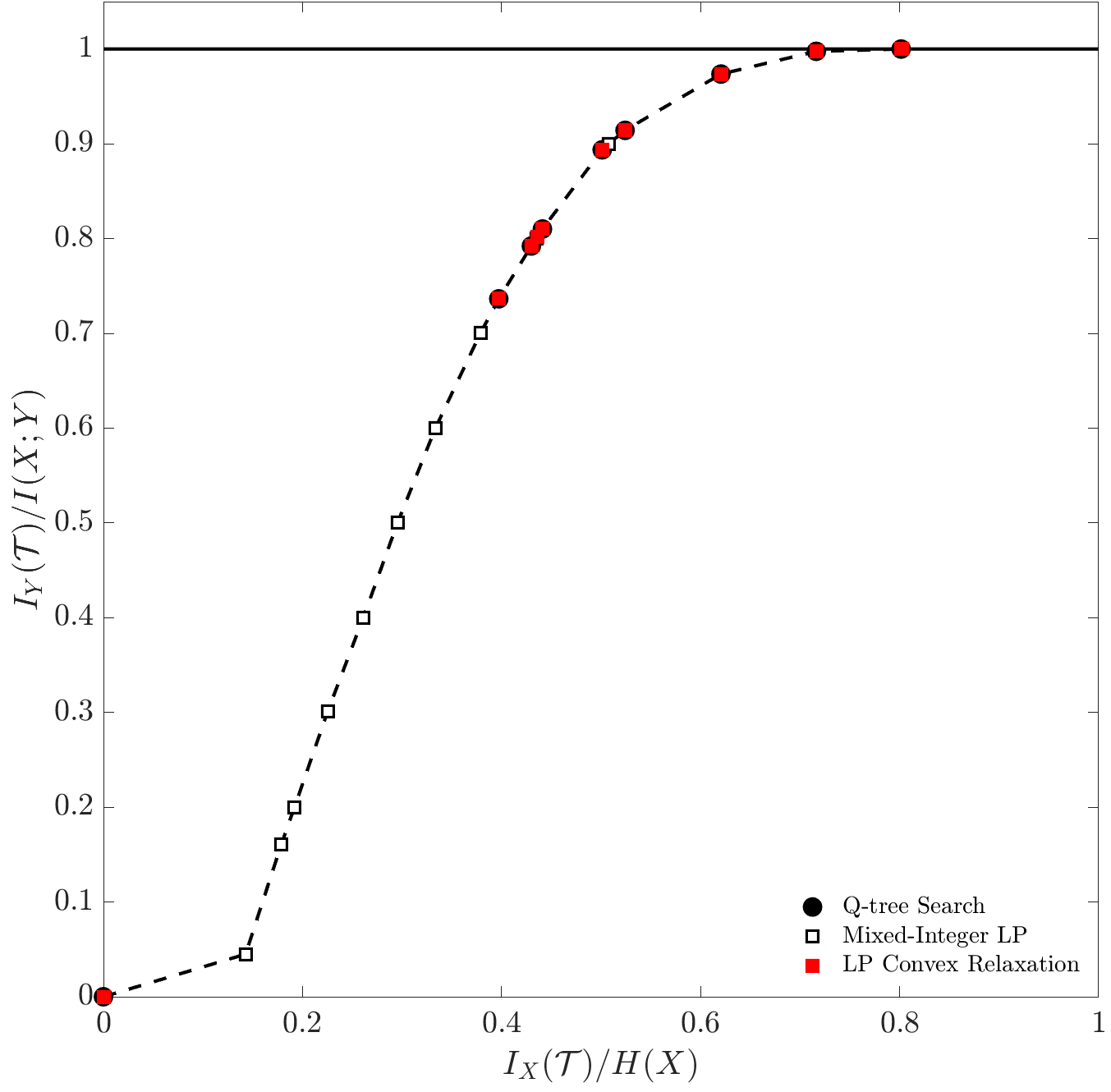}
	\caption{Normalized information-plane with solutions from Q-tree search, MILP and convex relaxation.}
	\label{fig:infoPlane}
\end{figure}

Shown in Figure \ref{fig:infoPlane} are solutions in the information-plane.
The information plane characterizes the trade-off between the amount of compression \(I(T;X)\) in comparison to the retained information \(I(T;Y)\) achieved by a compressed representation.
As the value of \(\hat D\) is increased, the solutions to \eqref{eq:intLinProg2}-\eqref{eq:intLinProg2_Cons3} are trees \(\T \in \T^\Q\) that retain more relevant information at the cost of less compression.
Thus, as \(\hat D\) (or \(D\)) is increased, one moves along the curve to the right.
For comparison, we provide the information-plane points obtained by executing the Q-tree search method from \cite{Larsson2020} as well those obtained by solving a convex relaxation of \eqref{eq:intLinProg2}-\eqref{eq:intLinProg2_Cons3}.
The convex relaxation results were obtained by replacing \eqref{eq:intLinProg2_Cons3} with \([\z]_t \in [0,1]\) for \(t \in \Nint(\T_\W)\), and truncating the returned CVX solution by considering those nodes for which \([\z]_t \geq 0.5\) as expanded.
By applying the same truncation criterion to all nodes \(t \in \Nint(\T_\W)\), we ensure via \eqref{eq:intLinProg2_Cons2} that the result is a feasible tree representation of \(\W\).

From Figure \ref{fig:infoPlane}, we see that the MILP returns solutions that are consistent with the Q-tree search method.
In addition, since the MILP formulation does not require guesswork to tune \(\beta\), we obtain a wider spectrum of solutions that characterize the trade-off between compression and information retention.
Additionally, the convex relaxation approach shows good agreement with both Q-tree search and the MILP formulation.
In fact, at the Q-tree search solutions shown, the three approaches provide equivalent solutions, and many times it is observed that the convex relaxation results in solutions that are equal to those obtained by the MILP.
Further investigation is needed in order to determine if by constructing a more sophisticated truncation rule the performance of the convex relaxation can be improved.

Lastly, note that the information-plane curve shown in Figure \ref{fig:infoPlane} is specific to abstractions in the form of multi-resolution quadtree structures.
As a result, it is not guaranteed that for a given value of \(\hat D\) there exist a solution \(\T \in \T^\Q\) to \eqref{eq:intLinProg2}-\eqref{eq:intLinProg2_Cons3} such that \(I_Y(\T) = \hat D\), as the space of quadtrees is discrete.
Consequently, the information-plane curve for encoders restricted to be feasible quadtree representations of \(\W\) is not a continuous function. 
This is because, for any tree \(\T \in \T^\Q\), the process of expanding an allowable node \(t \in \Nlf(\T)\) results in a finite, discrete, change in information, given by \(\Delta \hat I_X(t)\) and \(\Delta \hat I_Y(t)\) as a function of \(p(x,y)\).
This can be better understood by considering the visible ``kink" in the information-plane curve shown in Figure \ref{fig:infoPlane}.
Specifically, we see that, from a solution of \(I_X(\T) = I_Y(\T) = 0\) (i.e., full abstraction, corresponding to the tree \(\T = \mathcal{R}_\W\)), there is a jump in the information-plane curve to a value of \(\nicefrac{I_X(\T)}{I(X;Y)} = 0.14\) and \(\nicefrac{I_Y(\T)}{H(X)} = 0.04\).
This jump occurs because, from the root node/tree \(\mathcal R_\W \in \T^\Q\), there is only one allowable nodal expansion; namely to expand the root itself.
Once \(\hat D \geq \Delta \hat I_Y(t)\), where \(t = \mathcal R_\W\), the root will be expanded, resulting in the jump observed in Figure \ref{fig:infoPlane}.

\section{Conclusion}

In this paper, we developed an approach to obtain hierarchical abstractions for resource-constrained agents in the form of multi-resolution quadtrees.
The abstractions are not provided a priori, but instead emerge as a function of the information-processing capabilities of autonomous agents.
Our formulation leverages concepts from information-theoretic signal compression, specifically the information bottleneck method, to pose an optimal encoder problem over the space of multi-resolution tree representations.
We showed how our problem can be expressed as a mixed-integer linear program that can be solved using commercially available optimization toolboxes.
A non-trivial numerical example was presented to showcase the utility of our framework for generating abstractions for resource-limited agents. 
To conclude, we presented a discussion detailing the properties of our framework as well as a method for convex relaxation and a brief comparison of our approach to existing methods for abstraction.

\begin{acks}
	Support for this work has been provided by \grantsponsor{}{The Office of Naval Research}{} under awards \grantnum[]{}{N00014-18-1-2375} and \grantnum[]{}{N00014-18-1-2828} and by \grantsponsor{}{Army Research Laboratory}{} under \grantnum[]{}{DCIST CRA W911NF-17-2-018}.
\end{acks}

\bibliographystyle{ACM-Reference-Format}


\end{document}